\documentclass[11pt,a4paper]{article}
\usepackage{times,latexsym}
\usepackage{url}
\usepackage[T1]{fontenc}
\usepackage{hyperref}
\usepackage[acceptedWithA]{tacl2021v1}
\usepackage{graphicx} 
\usepackage{times}
\usepackage{latexsym}
\usepackage{booktabs}
\usepackage{enumitem}
\usepackage{multirow}
\usepackage{amssymb}
\usepackage{array}

\usepackage[utf8]{inputenc}
\usepackage{todonotes}

\usepackage{microtype}
\usepackage{float}
\usepackage[capitalise]{cleveref}
\usepackage[group-minimum-digits={3}]{siunitx}
\usepackage{colortbl}
\definecolor{lg}{gray}{0.89}

\newcommand{\longnum}[1]{\num[group-separator={,}]{#1}}
\title{Calibrated Interpretation: Confidence Estimation in Semantic Parsing}

\newcommand{\perc}[1]{$#1\%$}

\author{Elias Stengel-Eskin \\ Johns Hopkins University \\ {\tt{elias@jhu.edu}} \And Benjamin Van Durme \\
Johns Hopkins University \\
{\tt{vandurme@jhu.edu}}}

\begin{document}
\maketitle
\begin{abstract}
Sequence generation models are increasingly being used to translate natural language into programs, i.e. to perform executable semantic parsing. 
The fact that semantic parsing aims to predict programs that can lead to executed actions in the real world motivates developing safe systems. This in turn makes measuring calibration -- a central component to safety -- particularly important.
We investigate the calibration of popular generation models across four popular semantic parsing datasets, finding that it varies across models and datasets.  
We then analyze factors associated with calibration error and release new confidence-based challenge splits of two parsing datasets.
To facilitate the inclusion of calibration in semantic parsing evaluations, we release a library for computing calibration metrics.\footnote{Models/Analysis/Data: \url{https://github.com/esteng/calibration_miso}, Metric: \url{https://github.com/esteng/calibration_metric}}     
\end{abstract}

\defcitealias{semanticmachines2020}{SMCalFlow}
\defcitealias{cheng.j.2020}{TreeDST}
\defcitealias{yu.t.2018}{Spider}
\defcitealias{yu.t.2019}{CoSQL} 

\section{Introduction}
When probabilistic models are used for decision-making or interaction, we not only want the model to be accurate but also to be \emph{calibrated}, meaning that the probability the model assigns to a decision should roughly correspond to its likelihood of being correct. 
For example, if a model translates a user instruction into a program for a robot, the confidence that the program accurately captures the person's intent may inform whether the robot executes the program.
Because of its importance to safety and usability, a large body of work has addressed calibration, with most research considering single-timestep classification models (i.e. one output decision per input). 
However, probabilistic models are also commonly applied to multi-timestep  tasks. 
In these tasks, the model generates a \emph{sequence} of decisions, with each subsequent decision dependent on the input and the previous decisions.
For example, text generation models typically predict sequences of words one at a time, with each new word conditioned on the previous outputs. 
Although calibration has been measured in some text generation tasks like machine translation \citep{kumar.a.2019, wang.s.2020}, determining the accuracy of generated text sequences can be challenging, making calibration -- the relationship between accuracy and confidence -- hard to estimate.
As sequence generation models, especially large language models, play a role in a growing variety of tasks, measuring their calibration has become extremely important. 

The task of executable semantic parsing, where a model predicts an executable program from a natural language instruction, is often modeled as a sequence generation task. 
Such parsing models are used in systems that interact with the real world, such as human-robot interfaces \citep{tellex.s.2020} and digital assistants \citep{gupta.s.2018, cheng.j.2020, semanticmachines2020}.
Since actions in these domains -- especially physical domains -- can have irreversible effects, the importance of ensuring model safety cannot be understated. 
Thus, semantic parsing provides a clear motivation for having a well-calibrated sequence generation model:
At low confidence, we may prefer for the system to defer action or request clarification, while when confidence is high, these actions may unnecessarily annoy a user and make the system unusable.
This reasoning presupposes that the model's confidence is well-correlated with its probability of success. 

Simultaneously, the constrained and executable nature of semantic parses makes accuracy easier to measure.
In many text-based sequence generation tasks like machine translation, summarization, long-form question-answering (QA), and open-ended dialogue, evaluating the quality and correctness of a generation poses a variety of challenges, often due to the fact that there are many ways of stating roughly the same proposition in language.
Because calibration measures the relationship between accuracy and confidence, our ability to measure calibration will only be as good as our ability to measure accuracy.
In executable semantic parsing, the model generates a program (rather than text) with a more restricted vocabulary and known syntactic rules.
This generally limits the number of reasonable semantically equivalent outputs and makes quantifying accuracy easier.
Furthermore, the executable nature of the programs allows us to measure accuracy via denotation (i.e. the result of program execution) rather than form. 
These factors make it an ideal domain for benchmarking the calibration of sequence generation models.

In \cref{sec:calibration}, we conduct what is to our knowledge the first large-scale investigation of calibration in sequence generation models as applied to semantic parsing tasks. 
We examine a variety of commonly-used models and measure their calibration across four popular semantic parsing datasets drawn from two different domains: task-oriented dialogue (TOD) and text-to-SQL. 
We first document the model's calibration profiles, asking how well-calibrated modern semantic parsing systems are; this includes a large pre-trained model queried in a few-shot setting, as well as more traditional fine-tuned models. 
Using qualitative and quantitative metrics, we find that most TOD models are already fairly well-calibrated.
However, the same models are poorly-calibrated on text-to-SQL datasets. 
In \cref{sec:analysis}, we analyze various factors implicated in calibration, attempting to shed light on the differences documented in \cref{sec:calibration}.
We first find that dataset size does not account for the difference in calibration between TOD and SQL models.
Using measures of input and program difficulty, we then explore the relationship between difficulty and calibration, finding that for SQL programs, the difficulty of the input and output are associated with poor calibration. 
Finally, we find that on TOD datasets, each model's low-confidence examples are challenging for other models; this leads us to propose two confidence-based challenge datasets.
Because of calibration's importance to semantic parsing specifically -- and sequence generation tasks generally -- we introduce an open-source library for computing calibration metrics and plotting confidence, compatible with the Transformers library \citep{wolf.t.2020}.

\section{Related Work}
\subsection{Calibration in NLP}
Given the utility of calibrated models in decision-making, a large body of research has focused on describing the calibration characteristics of different architectures and models, with some work finding neural networks to be relatively well-calibrated \citep{niculescu.a.2005, minderer.m.2021, carrell.a.2022} -- including neural encoders pre-trained on text data \citep{desai.s.2020} -- and other research indicating they are not \citep{guo.c.2017, wang.s.2020, si.c.2022a}.
These mixed results preclude drawing general conclusions about neural models' calibration, and motivate studies like ours documenting the calibration characteristics of standard models. 

Past work has examined a variety of classification problems, often focusing on binary or multi-class classification \citep{naeini.m.2015, guo.c.2017, minderer.m.2021, khojah.r.2022}. 
Some papers have addressed sequential NLP tasks: e.g. \citet{jagannatha.a.2020} address calibration in structured prediction tasks.
More related to our sequence generation setting, \citet{kumar.a.2019} and \citet{wang.s.2020} examine calibration in machine translation, both finding models to be over-confident. 
Measuring calibration in translation tasks is limited by the metrics used, which are noisy proxies for accuracy and have well-documented limitations \citep{callison-burch.c.2006, mathur.n.2020}.
In semantic parsing specifically, past work has focused on improving confidence estimation for certain parsers.
\citet{dong.l.2018} propose a confidence estimation method based on model and input features.
Similarly, \citet{chen.s.2022} introduce an additional confidence-estimation model for semantic parses.
In our work, we extract confidence scores from the same model used for parsing, and focus on analyzing calibration rather than improving it. 

\begin{table*}
    \begin{tabular}{p{0.12\textwidth}p{0.06\textwidth}p{0.06\textwidth}p{0.06\textwidth}p{0.26\textwidth}p{0.29\textwidth}}
    \hline \hline
    Dataset &  Train  & Dev & Test & Example Input & Example Output \\
    \hline
    \citetalias{semanticmachines2020} & \longnum{108753} & \longnum{12271} &  \longnum{13496} & \emph{Do I have anything going on tonight?} & {\tt{(Yield (> (size (QueryEventResponse. results(...))) 0L))}} \\
    \hline
    \citetalias{cheng.j.2020} & \longnum{121652} & \longnum{22910} & \longnum{22841} & \emph{I want to book a flight to Paris} & {\tt{(plan (\^{}(Flight)... (Flight.dest...)))}} \\
    \hline 
    \citetalias{yu.t.2018} &  \longnum{7794} & \longnum{865} & \longnum{1034} & \emph{What are the numbers of all flights coming from Los Angeles?} & {\tt{SELECT flno FROM flight WHERE origin = ``Los Angeles''}}  \\
    \hline
    \citetalias{yu.t.2019} & \longnum{6598} & \longnum{745} & \longnum{1007} &  \emph{How many people are named Janessa? $|$ Do you mean the number of people whose first name is Janessa?  $|$  Yes} & {\tt{SELECT ...AS T1 JOIN ... T2.first\_name = ``Janessa''}} \\
    \hline
    \end{tabular}
    \vspace{-0.5em}
    \caption{Number of train, validation, and test examples per dataset and example inputs and outputs. For SQL tasks, column and table names are also included in the input (these are omitted here for readability.)}
    \label{tab:data}
    \vspace{-0.5em}
\end{table*}

\paragraph{Calibration in LLMs}
We focus on calibration in pre-trained large language models (LLMs), which has also been addressed in other lines of work, first by \citet{mielke.s.2022}, who examine calibration in large pre-trained dialogue models.
They focus on ``linguistic calibration,'' describing a guided generation method for introducing verbalized statements of uncertainty (e.g. ``I believe'', ``I'm not sure, but...'', ``I am certain...'', etc.) into QA responses. 
Our experiments use logit-based numerical confidence estimates instead. 
Similarly, \citet{lin.s.2022} analyze both logit-based and verbalized uncertainty in GPT-3. 
\citet{si.c.2023} examine LLM calibration through the lens of reliability, focusing on QA data and using logit-based confidence estimation. 
\citet{kadavath.s.2022} examine whether LLMs are well-calibrated by measuring the probability an answer is true.
Finally, \citet{zhou.k.2023} examine how teaching models to interpret and express certainty and uncertainty impacts calibration and performance. 
With the exception of \citet{kadavath.s.2022}, these studies focus on single-prediction classification settings: \citet{mielke.s.2022} examine TriviaQA questions \citep{joshi.m.2017}, \citet{lin.s.2022} use math questions, and both \citet{zhou.k.2023} and \citet{si.c.2023} consider multiple QA benchmarks including TriviaQA. 
While studying these settings is valuable, our focus is on longer-form sequence generation, perhaps a more typical use-case for LLMs. 

This divergence is also pointed out by \citet{kadavath.s.2022}, who, in addition to several experiments on QA benchmarks like MMLU \citep{hendrycks.d.2021}, include experiments on HumanEval code generation examples \citep{chen.m.2021} and their own dataset of Python code generation problems. 
Our experiments differ from theirs along a few axes. 
Firstly, although making claims about the calibration of ``language models'' broadly, \citeauthor{kadavath.s.2022} only consider a single pre-trained model (of varying sizes) and a single program synthesis task; we consider several models across four datasets.
Furthermore, while we consider both fine-tuned and few-shot models, \citeauthor{kadavath.s.2022} measure calibration only in a few-shot setting. 
While we obtain confidence estimates from the token probabilities, \citeauthor{kadavath.s.2022} extract their estimates via an additional prompt that asks the model to label a predicted program or answer as ``True'' or ``False'', where the confidence is taken to be $P(True)$. 
This is a natural formulation for few-shot models, but is less compatible with fine-tuned models, which are far more common in practice.
Note that \citeauthor{kadavath.s.2022}'s method incurs roughly twice the cost of program generation, as the generated program must be re-encoded to obtain a confidence estimate.

\section{Methods}
Our first goal is to determine the typical calibration characteristics of models applied to semantic parsing tasks. 
For this, we choose a range of standard semantic parsing tasks, datasets and models.

\subsection{Tasks and Datasets} 
In total, we examine four datasets across two tasks; statistics and examples of each are provided in \cref{tab:data}. 
All of the datasets we examine are in English. 
Two are task-oriented dialogue (TOD) parsing datasets: SMCalFlow \citep{semanticmachines2020} and TreeDST \citep{cheng.j.2020}. 

In the TOD task, a user engages in a dialogue with a digital assistant in order to achieve some goal, e.g. booking a flight or scheduling a meeting. 
For each user turn, the agent predicts an executable program and provides a response to the user based on the outcome of the program's execution; thus, the modeling task is to predict a program from a user input and a dialogue history. 
As neither SMCalFlow nor TreeDST has an available execution suite, we measure correctness by exact match to the reference program. 
For both TOD datasets, we use the preprocessed data from \citet{platanios.a.2020}, who converted TreeDST into a format shared with SMCalFlow that resembles Lisp, and use the SMCalFlow data splits given by  \citet{roy.s.2022}.

Our second task is database querying. 
As in TOD tasks, the modeling target is a program; however, the program here is a SQL query that can be executed against a database (DB) to return an answer. 
Crucially, this text-to-SQL task differs from the TOD task in that the programs are highly dependent on the structure of the execution environment. 
The DB's schema influences the program and changes depending on the DB being used. 
For the text-to-SQL task, we consider two popular datasets: Spider \citep{yu.t.2018} and CoSQL \citep{yu.t.2019}. 
Both datasets contain queries over a shared set of 200 DBs; however, while Spider queries are single-turn, CoSQL queries are multi-turn dialogues between a user and an agent. 
For SQL, we do have access to an execution engine, allowing us to measure both execution accuracy and exact-match accuracy. 
Here also, we use the splits and preprocessing scripts from \citet{roy.s.2022}. 

\subsection{Models} \label{sec:models}
Given our goal of benchmarking the calibration characteristics of different approaches commonly in use in semantic parsing, we measure calibration via two extant modeling paradigms: fine-tuning and in-context learning (ICL). 

\paragraph{Fine-tuning}
In the fine-tuning paradigm, we take a model pre-trained with a self-supervised objective on text or code data and continue to train it (i.e. fine-tune it) using a supervised objective on the training data for each dataset in \cref{tab:data}.\footnote{Excepting MISO, for which only part of text encoder is pre-trained and the rest of the model is trained from scratch.} 
We train seven models from two commonly-used semantic parsing paradigms: transductive and sequence-to-sequence (seq2seq). 
Transductive models \citep{zhang.s.2020} treat the parsing problem as a sequence-to-graph task. 
While the executable programs found in SMCalFlow and TreeDST are expressed as Lisp-like sequences, they also have an underlying execution graph. 
Rather than learning to generate the surface form, the transductive approach seeks to directly model the underlying graph, predicting a sequence of nodes as well as labeled, directed edges. 
\citet{zhang.s.2019a} and \citet{zhang.s.2019b} introduced the MISO transductive parsing framework for predicting directed acyclic graphs from text inputs.
MISO combines an encoder-decoder model for predicting nodes paired with a biaffine parser \citep{dozat.t.2016} for edge prediction. 
It features source and target copy operations allowing special tokens (such as names and numbers) to be copied from the input and previously generated tokens to be re-generated (in the case of re-entrancy in the execution graph).
This kind of model represents an ``engineering-heavy'' approach where inductive bias for parsing is encoded in the model architecture.
We take MISO as an example of this class, motivated by its application across a variety of semantic parsing tasks \citep{zhang.s.2019a, zhang.s.2019b, stengel-eskin.e.2020universal, stengel-eskin.e.2021tacl, stengel-eskin.e.2022smcalflow, li.z.2021calibrating}.
We use \citet{stengel-eskin.e.2022smcalflow}'s best model, which has a RoBERTa \citep{liu.y.2019} encoder and contains 127 million (M) parameters. 

The seq2seq paradigm instead directly models the output sequence.
While predicting the syntactic nuances of a parse (e.g. generating the correct number of closing parentheses) can be challenging, seq2seq models are better able to leverage large pre-trained Transformers, often enabling them to outperform methods with stronger inductive biases, like MISO. 
We use the BenchCLAMP framework \citep{roy.s.2022} to finetune seq2seq models for all TOD and SQL datasets; specifically, we examine the T5 \citep{raffel.c.2020} and BART \citep{lewis.m.2020} architectures, both of which have frequently been applied to semantic parsing benchmarks \citep{shaw.p.2021, scholak.t.2021, desai.s.2021, banerjee.d.2022}.
Both are large encoder-decoder Transformers \citep{vaswani.a.2017} pre-trained on text data with self-supervised objectives.
Since SQL is used in many open-source applications (unlike SMCalFlow and TreeDST), code scraped from the web is likely to contain examples of it; thus, for Spider and CoSQL, we also examine Code-T5 \citep{wang.y.2021}, a T5 architecture pre-trained on large amounts of web-scraped code.
As MISO is not commonly used in text-to-SQL tasks, we choose to omit it in our SQL experiments.
We examine T5-small (60M parameters), T5-base (220M), T5-large (770M), BART-base (139M), and BART-large (406M), as well as Code-T5-base (220M).  
Note that while BenchCLAMP allows for constrained decoding according to a context-free grammar, restricting the model to producing only valid parses, we choose to decode in an unconstrained fashion. 
Because the constrained decoding process intervenes on the output logit space, zeroing out invalid continuations and renormalizing, it could affect the model's calibration characteristics. 

\paragraph{In-Context Learning}
The ICL paradigm instead uses a model that has \emph{only} been trained with a self-supervised objective.
\citet{brown.t.2020} find that a sufficiently large model pre-trained on text data can perform many tasks after being shown a few examples, without any updates to the gradients. 
This has been extensively explored in the semantic parsing domain, where grammar-constrained decoding has been combined with ICL, allowing models to predict dataset-specific programs from only a few retrieved examples \citep{shin.r.2021, shin.r.2022, roy.s.2022}.
Following this paradigm, for a given test query, we retrieve a set of relevant input-output examples from the training data and concatenate them into the model's input, or prompt, as instructive examples.
The model then predicts a parse, constrained by the grammar of the domain language.
Past work in ICL for semantic parsing has used OpenAI's Codex model \citep{chen.m.2021}, a LLM based on the GPT architecture and trained on web-scraped text and code.
However, in addition to being costly, Codex was recently slated for removal from the OpenAI's public API. 
To ensure our results are not affected by similar decision in the future, we instead opt to use open-source publicly-released models for our ICL experiments.
We use Codegen \citep{nijkamp.n.2022}, which is available through the  Transformers library \citep{wolf.t.2020}.
Codegen models are also auto-regressive GPT-style models pre-trained on code, with strong performance across code generation tasks; we explore 4 model sizes: 350M parameters, 2B, 6B, and 16B.

For each test example, we construct a prompt by retrieving 5 similar training examples and concatenating them into the context.\footnote{The number of examples was chosen in light of hardware memory limitations when running the largest models.
Similarly, all billion-parameter models were run at half precision, and memory constraints precluded running constrained decoding on the 16B model, so we do not provide sequence-level accuracies for this model.}
We follow \citet{roy.s.2022} and use a BM25 retriever \citep{robertson.s.2009} over the full training set, indexed by the input utterance. 
Examples are ordered by similarity, with the most similar example appearing last (i.e. immediately before the test input). 

\paragraph{Input representation}
In TOD datasets like SMCalFlow and TreeDST, the user communicates with the agent over the course of a dialogue; 
we follow previous work in using the previous dialogue turn as input if available. 
Thus, each datapoint consists of an input $X = (\mathcal{U}_0, \mathcal{A}_0, \mathcal{U}_1)$ and an output program $\mathcal{P}$, where $\mathcal{U}_0$ is the previous user utterance (if it exists), $\mathcal{A}_0$ is an automatically-generated agent response to the previous utterance, and $\mathcal{U}_1$ is the current user utterance. 
Similarly, for CoSQL we include the previous dialogue turn in the input (if available). 
Correctly predicting a SQL program relies on knowledge of the DB schema; to inform the model of the schema, we follow past work in concatenating schema information (column and table names) into the input. 

\paragraph{Confidence Estimation}
The models used (except for MISO) predict subword tokens, rather than the whitespace-delimited tokens of a programming language. 
For example, the SQL token {\tt{SELECT}} is split into 3 subwords by the BART tokenizer; in other words, the model will produce 3 probabilities for this single program token. 
We estimate program token confidence by first obtaining estimates for subwords and then aggregating. 
When measuring token-level accuracy, we also use program tokens rather than subwords.

To estimate subword-level confidence, we use the baseline estimator introduced by \citet{hendrycks.d.2016}, which is robust across many tasks \citep{varshney.n.2022}.
More specifically, we take the maximum probability across the output vocabulary at each timestep. 
We then aggregate the probabilities across subwords of a given token using {\tt{min}}.
Intuitively, a prediction is only as good as its weakest link. 
Given a sufficiently long sequence of subwords, aggregation methods like {\tt{mean}} may obscure low-confidence decisions. 
In practice, using {\tt{mean}} pooling on subwords gives qualitatively similar results, as the number of subwords per whitespace token tends to be small.
However, in \cref{sec:sequence} we see that {\tt{mean}} leads to high ECE on sequence-level calibration, indicating that {\tt{min}} may be the better overall choice.  

\begin{figure*}[ht]
    \centering
    \includegraphics[width=\textwidth]{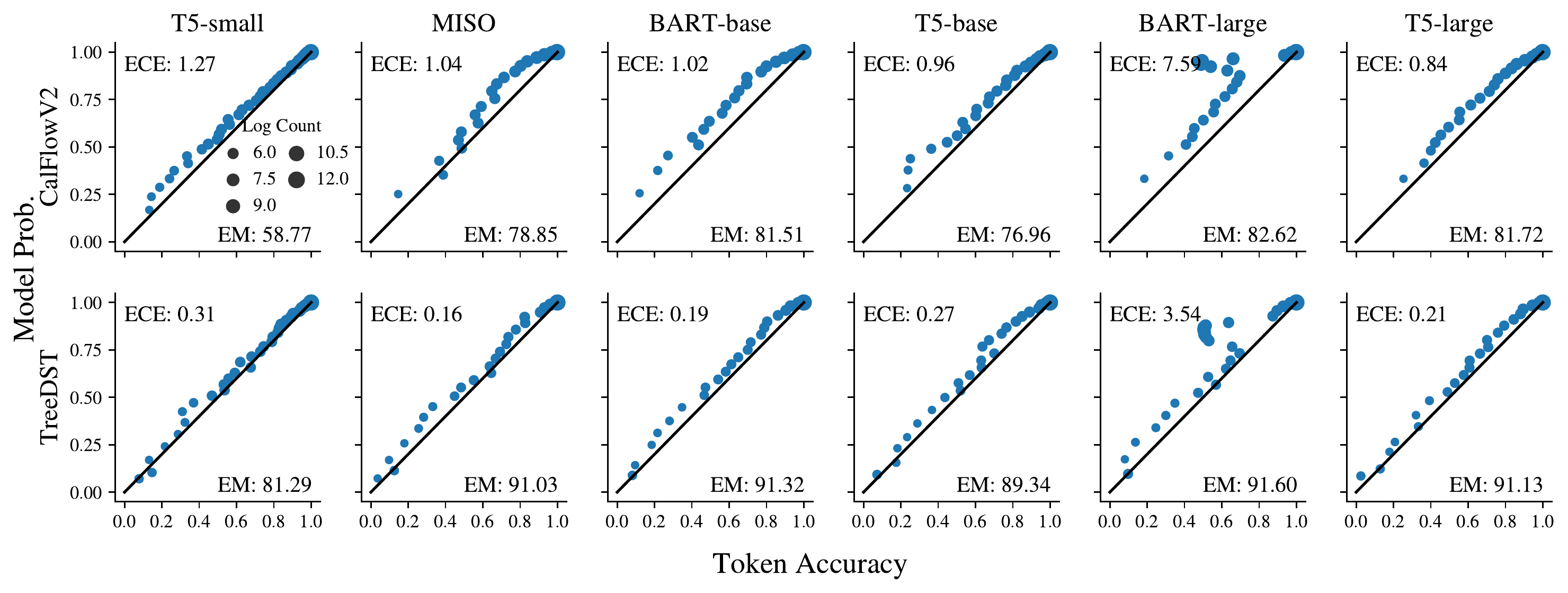}
    \vspace{-2em}
    \caption{Token-level model confidence and mean accuracy, binned by confidence across models (sorted by size) for TOD datasets. Point size reflects the number of tokens in the bin. Points above the line reflect overconfidence, while those below reflect underconfidence. Exact Match accuracy (EM, higher is better) and Expected Calibration Error (ECE, lower is better) are given. All models show relatively low ECE.}
    \label{fig:all_models}
    \vspace{-0.5em}
\end{figure*}

\subsection{Metrics} 
For TOD and SQL datasets, we evaluate our models' semantic parsing ability using exact match accuracy (EM), where a prediction is considered correct it it exactly matches the reference program.
This can be quite strict and lead to false negatives; for example, the snippet {\tt{x > 0 AND x < 5}} is logically identical to the snippet {\tt{x < 5 AND x > 0}} but would result in an EM score of 0. 
In the case of SQL, we could mitigate this by executing programs against the provided DBs and comparing the result of the execution to the reference result. 
This metric is perhaps too lenient, and can result in false positives, e.g. if the gold program yields a null result, then any program yielding a null result will be counted as correct, even if it does not correspond at all to the user's input. 
In light of this, \citet{zhong.r.2020} introduce test-suite accuracy, which executes each program against a suite of test DBs optimized for high code coverage.
A program is counted as correct if it matches the gold program's denotations on all DBs in the suite.
This reduces the false positive rate and provides a much tighter upper-bound on performance.

We follow past work in using expected calibration error (ECE) \citep{naeini.m.2015, guo.c.2017} as our calibration metric.
To compute ECE, predictions are binned by the model's confidence;  since each prediction has an accuracy of either 0 or 1, accuracies must be averaged across examples falling in a confidence range to obtain an expected accuracy for a given confidence score. 
In its original formulation, the number of bins used is a hyperparameter. 
However, \citet{ding.y.2020} find that a fixed binning approach used in ECE can be suboptimal. 
Intuitively, because confidence scores are often distributed non-uniformly, a fixed binning strategy may result in bins containing few samples, leading to high-variance accuracy estimates within these bins. 
In general, there is a tradeoff between introducing many small bins (estimating accuracy with high variance) and maintaining a few large bins (estimating accuracy with high bias). 
They instead introduce \emph{adaptive binning}, which correlates the number of samples in a bin with the bin's range: in regions where confidence estimates are sparse, adaptive binning introduces more bins, such that a smaller range is covered by each bin (reducing bias), while when estimates are dense, adaptive binning includes fewer bins, reducing the variance of the estimate. 
The number of samples for each bin is given by $n = 0.25\big(\frac{Z_{\alpha}/2}{\epsilon}\big)^2$, where $Z_{\alpha/2}$ is the standard normal distribution's Z-score, $1-\alpha$ is the confidence interval, and $\epsilon$ is a small positive value included for numerical stability.
\footnote{Like \citet{ding.y.2020}, we found that the metric is not sensitive changing hyperparameters.}

The expected calibration error is the difference between the average accuracy of each bin and its confidence, weighted by the size of the bin. 
Let $\hat{\mathcal{Z}}$ be the model's distribution over the output vocabulary $\mathcal{V}$, and let $\hat{C} = \max \hat{\mathcal{Z}}$, $\hat{Y} = \text{argmax} \:\hat{\mathcal{Z}}$.
Let $Y$ be the true class indices and define a binary accuracy vector $A$ s.t. $a_i = \delta(\hat{y}_i, y_i)$.
After binning $\hat{Y}$ into $N$ bins $\mathcal{B}$, $ECE(\mathcal{B})$ is defined as:
\begin{equation}
    ECE(\mathcal{B}) = 100 * \sum\limits_{i=1}^{N} \frac{|\mathcal{B}_i|}{N} \Big| \frac{\sum\limits_{j\in \mathcal{B}_i} a_j }{|\mathcal{B}_i|} - \frac{ \sum\limits_{j\in \mathcal{B}_i} c_j}{|\mathcal{B}_i|}\Big|
    \label{eqn:ece}
\end{equation}
In other words, ECE is the mean absolute error between each bin's average confidence and average accuracy; we scale ECE by a factor of 100 for readability. 
In addition to ECE, we qualitatively analyze calibration by plotting the average accuracy against the bin confidence.
To encourage calibration to be measured in semantic parsing, we release our metric and plotting library as a Python package. 

\begin{figure*}[ht]
    \centering
    \includegraphics[width=\textwidth]{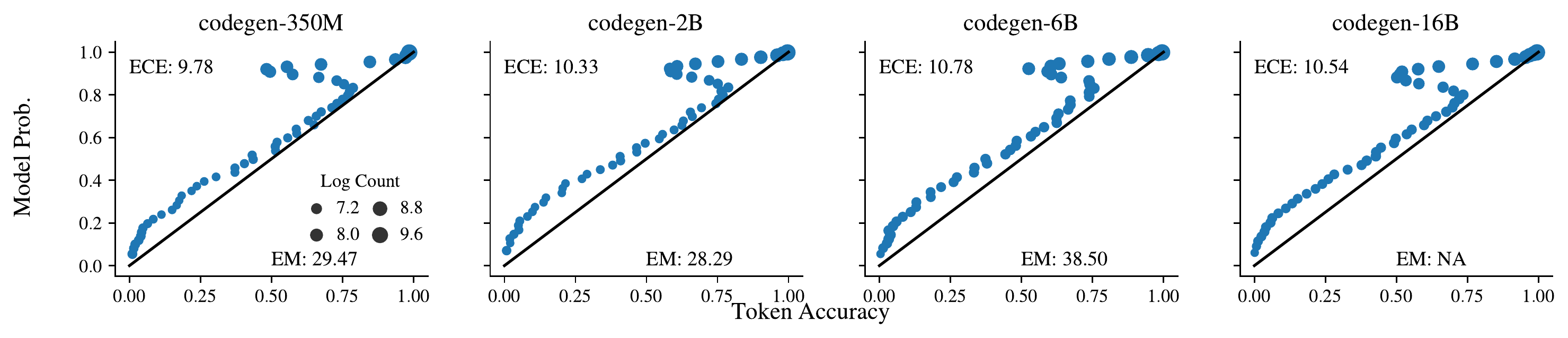} 
    \vspace{-2em}
    \caption{Token-level calibration curves for few-shot code models tested on SMCalFlow.}
    \label{fig:few_shot}
\end{figure*}

\section{Benchmarking Calibration} \label{sec:calibration}
Calibration can be measured at the token-level or the sequence-level.
Sequence-level calibration is most relevant to safety; in a system using an executable parsing model, the predicted program will be executed as a whole (i.e. as a sequence). 
However, absent a (potentially expensive) external model, sequence-level confidence scores will be composed of token-level scores, just as sequences are composed of tokens. 
Moreover, token-level scores can reveal phenomena obscured by sequence-level scores.
Rather than only measuring the model's accuracy and confidence on a whole sequence, it may be more informative to examine on which types of tokens the model makes mistakes or has high calibration error.
For example, token-level confidence scores can allow us to examine which specific functions the model is struggling on.

\begin{figure*}[t]
    \centering
    \includegraphics[width=\textwidth]{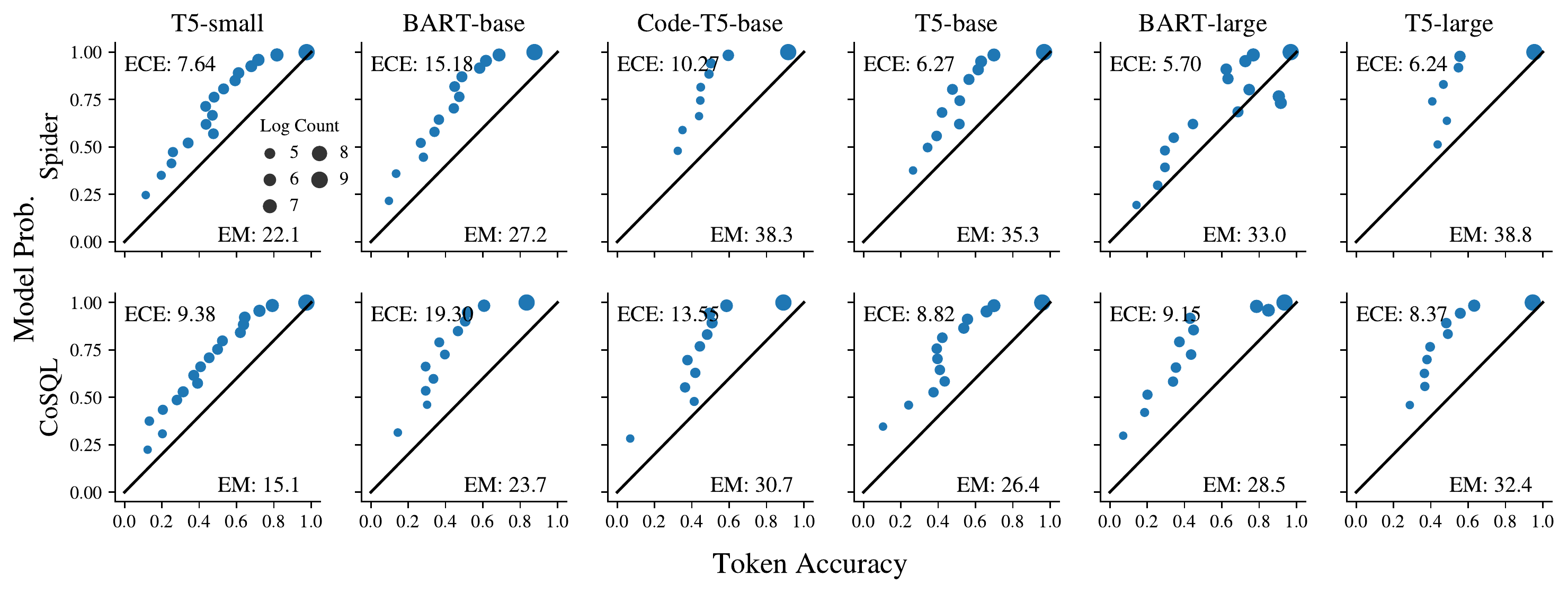}
    \vspace{-2em}
    \caption{Token-level model confidence and accuracy for SQL datasets, binned by confidence across models and datasets. Unlike the models fine-tuned on TOD data (cf. \cref{fig:all_models}) models fine-tuned on SQL are poorly-calibrated.}
    \label{fig:all_models_sql}
    \vspace{-1em}
\end{figure*}

To measure token-level accuracy against a reference program at time $t$, all predicted tokens at timesteps $1,\ldots,t-1$ must match the reference program's prefix; thus, we use teacher forcing, feeding the model the gold prefix up to the current timestep. 
I.e. when predicting the confidence of token $\hat{y}_t$, we use tokens $y_1, \ldots, y_{t-1}$ taken from the \emph{gold} program $P$, not the predicted program $\hat{P}$.
Note that this differs from the setting used for computing EM accuracy in \cref{fig:all_models}, \cref{fig:few_shot}, and \cref{fig:all_models_sql}, where we compare the gold program $P = y_1, \ldots, y_T$ to the predicted program $\hat{P} = \hat{y}_1, \ldots, \hat{y}_T$. 
For sequence-level confidence, we use $\hat{P}$.

\subsection{Task-oriented Dialogue Results} \label{sec:tod_results}
\cref{fig:all_models} shows the token-level calibration plots and ECE scores for all fine-tuned models on TOD datasets. 
Note that in our plots, the model's confidence is shown on the y-axis, and the accuracy on the x-axis. 
While this is non-standard, it leads to a more natural interpretation of the plot w.r.t. the line $y=x$: points above the line are overconfident bins (confidence $>$ accuracy); those under the line are underconfident. 
The models are ranked by size.
The size of each point is based on the log of the number of elements in that bin (following \citet{mielke.s.2022}); the largest bin for all models is the most confident bin. 
This is consistent with the Exact Match (EM) accuracy results reported in \cref{fig:all_models}; for a model to achieve high accuracy, all its output tokens must exactly match the reference tokens on most programs, i.e. the vast majority of tokens must be predicted correctly. 

\begin{figure}[H]
    \centering
    \includegraphics[width=0.5\textwidth]{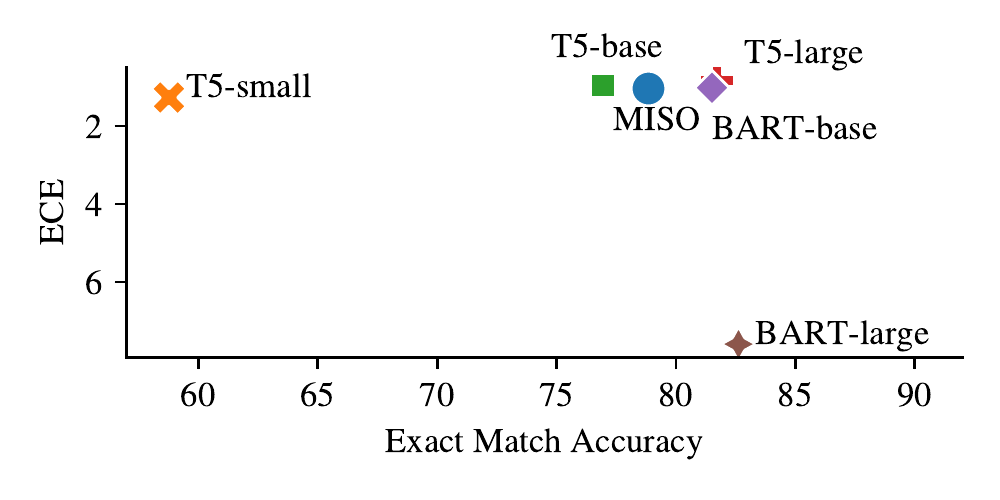}
    \vspace{-2em}
    \caption{Empirical trade-off between accuracy and calibration.
    Note that the Y-axis is flipped (lower calibration error is better). 
    Several models exist at the Pareto front.}
    \label{fig:pareto}
\end{figure}

We note that all models are relatively well-calibrated.
For the T5 series, calibration improves with scale, while the opposite is true for BART. 
In \cref{fig:pareto}, we plot the trade-off between accuracy and calibration error for SMCalFlow models on the test set. 
Here, we observe a Pareto front, with BART models having higher accuracy but worse ECE than comparable T5 models. 
Importantly, BART's calibration curves are non-monotonic, which is particularly troubling. 
Given monotonic distortions (e.g. a sigmoid calibration curve), regression models can be fit to the validation set to correct the calibration curve \citep{zadrozny.b.2002}.
Such correction is much harder to do for non-monotonic distortions.

\subsection{Few-shot Calibration}

\cref{fig:few_shot} shows the calibration curves for ICL models with 5 prompt examples on the SMCalFlow test set. 
First, we note that the exact-match results here are obtained with constrained decoding, since without fine-tuning models are generally unable to produce syntactically-correct programs. 
However, the token-level confidence scores are obtained without constraints (as in \cref{sec:tod_results}), so the curves are comparable to those in \cref{fig:all_models}. 
Even with constraints, the exact match performance of these models is lower than that of the fine-tuned models; however, they are shown only five examples per input, while the fine-tuned models are trained on over \longnum{100000} examples. 
The ECE of the ICL models is higher than the ECE of finetuned models. 
While ECE increases initially with model scale, the ECE of the 16B model is slightly lower than the ECE of the 6B model, suggesting that the trend could be U-shaped, a phenomenon that has been documented in other large models \citep{wei.j.2022}. 
Qualitatively and quantitatively, the ICL calibration curves are remarkably similar to the BART-large curve in \cref{fig:all_models}, despite the fact that BART-large is fine-tuned on the training data. 
We qualitatively analyze the spikes seen in many BPE-based models like BART and Codegen in \cref{sec:qualitative}, finding that they are driven by common syntactic tokens.

\subsection{SQL Results}
\cref{fig:all_models_sql} shows the token-level calibration characteristics of models fine-tuned on SQL data.
Adaptive binning results in fewer bins here due to the smaller sizes of the SQL test sets. 
We see both qualitatively and quantitatively that many of the same models which were well-calibrated on TOD data are poorly-calibrated on SQL datasets.
All models are substantially over-confident, and the trends between models here are different than in the TOD setting.
BART-large is better-calibrated than BART-base, which is the worst-calibrated on both datasets. 
We also see each model has higher ECE on CoSQL than Spider; this may have to do with input complexity, since CoSQL is multi-turn. 
We also note that the EM scores for SQL are generally quite low; this tracks with past results finding that EM is an excessively strict metric for SQL \citep{zhong.r.2020}.
We found the execution accuracies for the models to be in-line with those reported by \citet{roy.s.2022}

\subsection{Sequence-level Calibration} \label{sec:sequence}
The results in \cref{fig:all_models} follow past calibration work in using confidence estimates at the level of individual classifications (i.e. tokens). 
However, unlike many of the NLP tasks where calibration has been explored in the past, the output in semantic parsing is sequential, i.e. a series of classifications, where each timestep is dependent on the preceding decisions. 
Thus, to obtain sequence-level confidence scores from token-level scores, we need a method for aggregating token-level estimates.  
We explore two aggregation functions: {\tt{min}} and {\tt{mean}}. 
These operate over the token-level confidence scores generated during decoding; note here the decoding takes place \emph{without} teacher-forcing, i.e. using the predicted program $\hat{P}$. 
\cref{tab:seq_level} shows the ECE for sequence-level confidence scores with the two aggregation methods on the SMCalFlow dataset.
Since {\tt{min}} is typically better, especially on BART-large and T5-large, where {\tt{mean}} results in high over-confidence, we adopt {\tt{min}} moving forward. 
As mentioned in \cref{sec:models}, {\tt{mean}} may result in high ECE because over a long sequence, a single low-confidence prediction may be ``washed out'' by a sequence of high-confidence predictions. 
This is especially true in program prediction tasks, where syntactic constraints mean that many tokens can be easily predicted with high confidence (e.g. most SQL programs begin with {\tt{SELECT}}, etc.). 

\begin{table}[ht]
    \centering
    \begin{tabular}{lcc}
    \hline\hline
    Model & ECE (Min.) & ECE (Mean) \\
    \hline\hline
    MISO &  5.57 & 5.49 \\
    BART-large &  6.23 & 16.85 \\
    T5-large & 8.29 & 18.01\\
    \hline 
    \end{tabular}
    \caption{Sequence-level ECE for representative models on SMCalFlow.}
    \label{tab:seq_level}
    \vspace{-0.5em}
\end{table}

\section{Discussion and Analysis} \label{sec:analysis}
In \cref{sec:calibration}, we found that for a fixed dataset, different models have varying calibration characteristics, even when the models have similar architectures and pre-training data. 
Similarly, we saw that the same models often differ between datasets, and especially differ between domains (TOD vs. text-to-SQL). 
Finally, we found that some models show conspicuous spikes in calibration error in some regions of the confidence space, and that these spikes hold across models.
In light of these differences and quirks, we conduct additional analyses on factors associated with model calibration. 

\subsection{Qualitative Analysis} \label{sec:qualitative} 

\paragraph{Error spikes} 
In \cref{fig:all_models} (for BART-large) and in \cref{fig:few_shot} for all models, we see a spike in calibration error in the higher-confidence region.
This spike generally becomes more pronounced as the size of the model increases. 
Given that these models all share a tokenizer and that the spikes are located in similar regions of the confidence space, we hypothesize that they might be related to the identity of the tokens in the bins.
We isolate predictions in bins within each spike (i.e. bins with high confidence and lower accuracy, where the curve deviates from the $y=x$ line.)
We find that many of the tokens are shared between models and datasets; while each bin for each dataset does contain ``content'' tokens (e.g. function names, value names), all bins for all models contain a large number of very frequent ``syntactic'' tokens (e.g. {\tt{(,),=,>}}, etc.) shared by SMCalFlow and TreeDST.
Given how common these tokens -- especially parentheses -- are in the Lisp datasets, it is likely that the error spikes are driven by overconfidence on these syntactic tokens. 

\paragraph{Common errors} 
We find that most models are over-confident; we can see over-confidence either as a failure of confidence estimation (i.e. the model predicts a higher confidence value than is appropriate) or a failure of prediction (i.e. the model makes mistakes that lead to low accuracy). 
Following the second interpretation, we examine some common errors that models make on text-to-SQL and TOD datasets. 
For SQL datasets, we find that higher-confidence programs ($>0.5$) sometimes are semantically correct, but fail according to exact match because of a syntactic difference. 
For example, some programs use single quotes rather than double quotes; while this does not affect the execution result, it counts as a failure for exact-match accuracy. 
We discuss the shortcomings of exact-match further in \cref{sec:execution}. 
In SMCalFlow, many high-confidence errors are due to mismatches in capitalization, and others are due to confusion between similar functions, e.g. substituting {\tt{AttendeeListHasRecipientConstraint}} for {\tt{AttendeeListHasRecipient}}. 

We also see some errors due to ambiguous examples, where the model predicts a valid interpretation of the input which deviates from the reference. 
For example, given the input \emph{``List the number of different series names and contents in the TV Channel table.''}, the BART-large text-to-SQL model trained on Spider predicts the program:

\noindent{\tt{SELECT count ( DISTINCT series\_name ) , content FROM tv\_channel}}

\noindent which corresponds to the following valid parse of the input: 
\emph{``[the number of different series names] and [contents]...''}.
The reference however is:

\noindent{\tt{SELECT count ( DISTINCT series\_name ) , count ( DISTINCT content ) FROM tv\_channel;}}

\noindent which corresponds to \emph{``[the number of different [series names and contents]]..."}.

\subsection{Data size} \label{sec:data_size} 
Given the dramatic differences in calibration between \cref{fig:all_models} and \cref{fig:all_models_sql}, a natural question to ask is \emph{why} SQL models are so much less calibrated.
One simple hypothesis is dataset size:
\cref{tab:data} shows that the SQL datasets have between 14x and 18x \emph{less} data available for fine-tuning than the TOD datasets. 
Could simply increasing the size of the dataset fix the poor SQL calibration seen in \cref{fig:all_models_sql}?

If calibration were a result primarily of the dataset size, we would expect an SMCalFlow model trained on a small amount of data to also have high ECE. 
To test this, we trained a T5-small model on the first \longnum{7794} examples in the SMCalFlow dataset -- the same exact number of training examples as are available for Spider.
We find that the model's ECE increases from 1.27 to 2.40. 
While the ECE does increase, it is still far lower than the ECE on Spider (7.64), indicating that low dataset size is not the primary driver of poor calibration on text-to-SQL tasks. 

\subsection{Input and Output Difficulty} \label{sec:perplexity} 
Having established that the difference in calibration between TOD and text-to-SQL models cannot be accounted for merely by dataset size, we explore other factors that could be associated with poor calibration. 
Specifically, we examine the difficulty of the input and output. 
For the input, we examine out-of-distribution (OOD) inputs, following past observations that models are typically over-confident on OOD inputs \citep{bui.h.2023}.
We measure how OOD an input is via perplexity from an LSTM LM \citep{hochreiter.s.1997}.
To exclude the possibility of data leakage, we train this model from scratch, without any pre-trained word embeddings.
Inputs are tokenized using BPE \citep{sennrich.r.2016} and the train split is used to build a vocabulary.
We autoregressively train 2-layer unidirectional 256-dimensional LSTMs on the training splits of SMCalFlow  and Spider, using an Adam optimizer \citep{kingma.d.2014} with a learning rate of $0.001$.\footnote{LSTMs were chosen over Transformers here because of the relatively small size of the data combined with the desire to train the model from scratch.} 
Training ends when the validation perplexity fails to decrease for 5 consecutive epochs.

\begin{figure}[ht]
    \centering
    \includegraphics[width=0.5\textwidth]{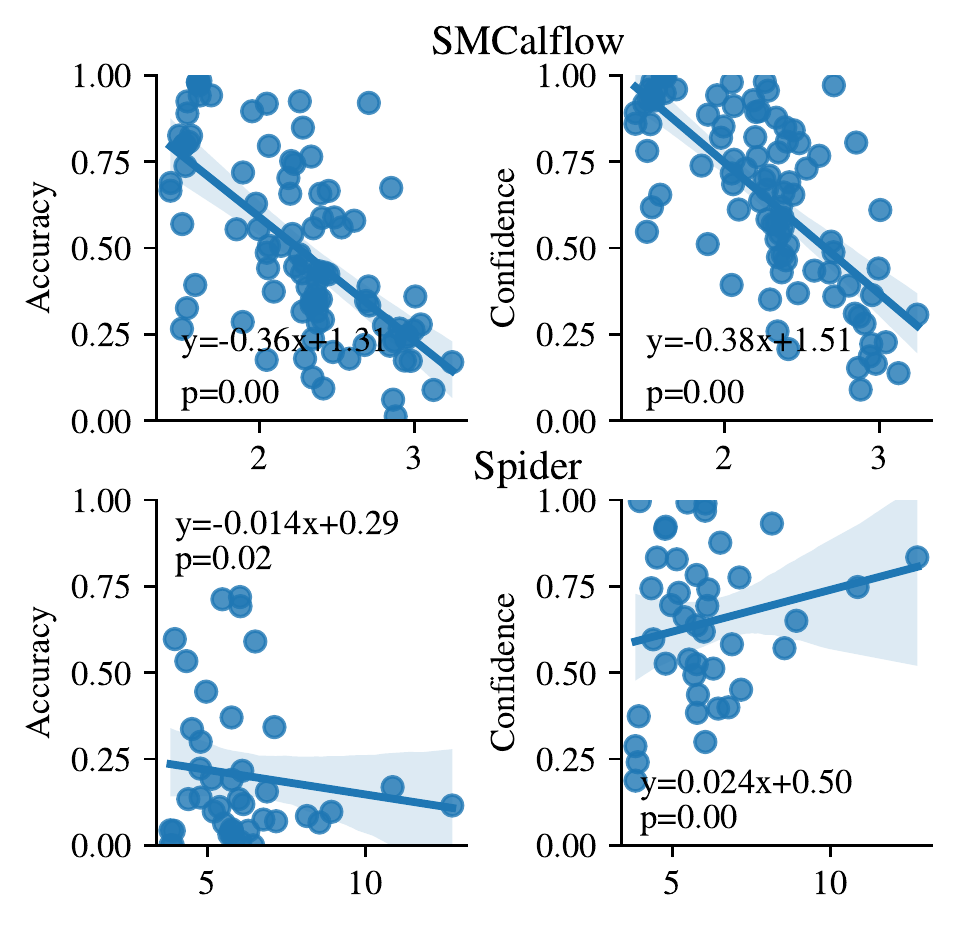}
    \vspace{-2em}
    \caption{Confidence and accuracy (y-axis) plotted by perplexity of the input (x-axis).}
    \label{fig:perplexity}
    \vspace{-0.5em}
\end{figure}

We consider sequence-level confidence bins from models shared between SMCalFlow and Spider (this excludes MISO and Code-T5). 
We compute the average perplexity of the inputs in each bin and plot the confidence and accuracy against the mean perplexity in \cref{fig:perplexity}. 
In all cases, we expect to see accuracy decreasing with perplexity: intuitively, as inputs become more OOD, they become harder for the model to parse. 
If models are robust to OOD inputs, the confidence should also decrease with perplexity, i.e. the model should be ``aware'' of the fact that it is worse at predicting programs for OOD inputs. 
On SMCalFlow, accuracy is correlated with input perplexity, decreasing on more OOD inputs. 
Confidence decreases with perplexity at a similar rate, indicating that the model has learned to identify OOD inputs and produce lower confidence values accordingly. 
That the slopes of the two lines are very close aligns with the low ECE seen on SMCalFlow across models: accuracy, confidence, and input perplexity are all ``coupled''.

For Spider, on the other hand, there is no such coupling. 
While accuracy is weakly correlated with perplexity, confidence is \emph{positively} correlated with perplexity. 
In other words, as inputs become more OOD, the model becomes \emph{more} over-confident in its predictions.\footnote{Note that for Spider, the two outlier points with high average perplexity ($>10$) contain many tokens not seen in training, like names of countries. 
Removing these outliers does not have a qualitative impact on the results.}
This suggests that the SQL models may not have learned to recognize OOD inputs.

\begin{figure}
    \centering
    \includegraphics[width=0.5\textwidth]{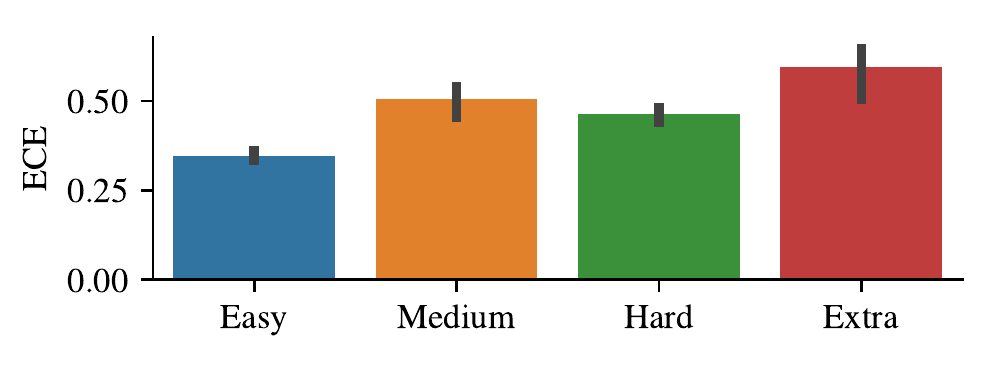}
    \vspace{-2.5em}
    \caption{T5-large ECE by target program difficulty on Spider (as defined by \citet{yu.t.2018})} 
    \vspace{-0.5em}
    \label{fig:tgt_diff}
\end{figure}

We also consider output difficulty in the SQL domain, where use the difficulty labels provided by \citet{yu.t.2018}, who classify SQL programs into \emph{easy}, \emph{medium}, \emph{hard}, and \emph{extra-hard} depending on the types of functions they involve.
We use T5-large, which is the best-calibrated (and best-performing) model in \cref{fig:all_models_sql}, and compute the sequence-level ECE for programs separated out by difficulty-type. 
We do note here that, due to the relatively small number of programs in the SQL test set, the bins here are often sparse, introducing variance into the estimates. 
To account for this, we train and evaluate models using 3 random seeds and plot the average performance. 
These results are described in \cref{fig:tgt_diff}, where ECE generally increases with difficulty.
The extra-hard programs have roughly twice the ECE of easy programs.
Notably, accuracy decreases as programs get harder; the high ECE values for harder programs indicates that the model does not lower the confidence appropriately for these programs. 

\subsection{Execution Accuracy} \label{sec:execution}
Because SQL is executable, we can additionally examine whether models are well-calibrated with respect to execution accuracy. 
This is especially relevant in light of the results in \cref{fig:all_models_sql}, indicating that models are over-confident.
Execution accuracy is typically more lenient than exact match accuracy, which is prone to \emph{false negatives}; a program may vary syntactically from a reference but still execute to the same result, i.e. have the same denotation. 

Another more lenient form of accuracy is accuracy@$k$ (Acc@$k$, which measures whether the correct program is in the top $k$ programs returned by the model after beam search. 
Intuitively, if there are two equally valid programs for a given input, we would expect Acc@$2$ to capture both, while Acc@$1$ (i.e. standard EM accuracy) would only have a $50\%$ chance of being correct.
While Acc@$k$ is more lenient, it is not a realistic metric, since in practice we always need to choose one single program to execute. 

\begin{figure}
    \centering
    \includegraphics[width=0.5\textwidth]{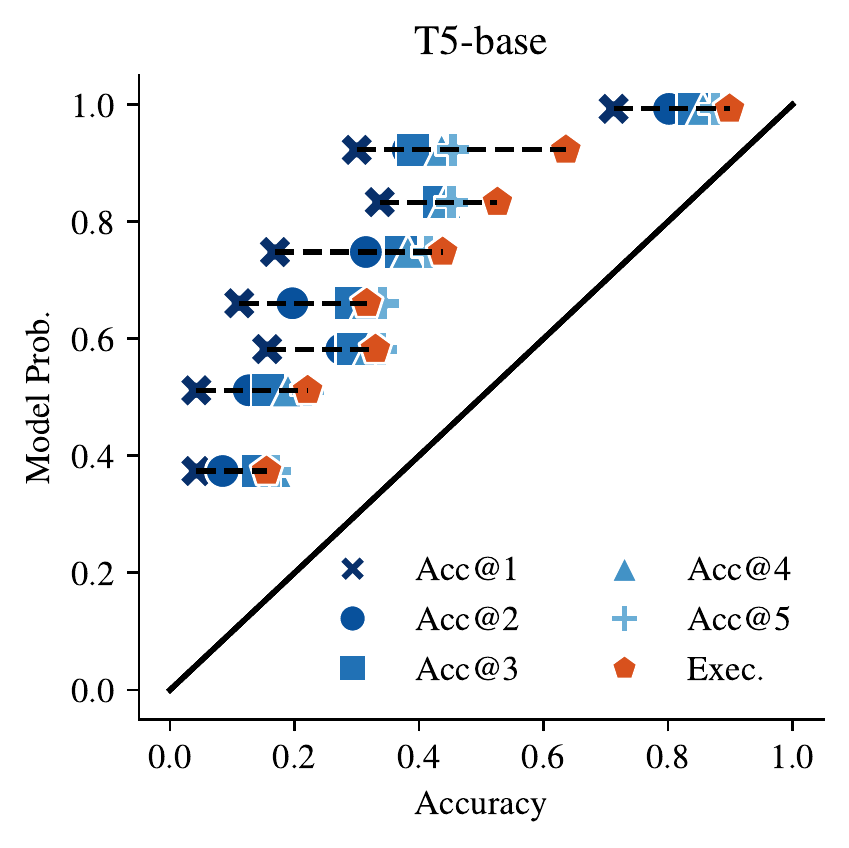} 
    \vspace{-2.5em}
    \caption{Execution accuracy and exact match accuracy@$k$. All accuracies result in over-confidence, but execution accuracy is generally the best-calibrated.}
    \vspace{-0.5em}
    \label{fig:exec_acc}
\end{figure}

\cref{fig:exec_acc} shows Acc@$k$ and the execution accuracy for T5-base on the Spider test set. 
The dotted line connects Acc@$1$ to execution accuracy, showing the range. 
First, we note that while all bins are over-confident, the execution accuracy is generally better-calibrated (less over-confident) than any of the Acc@$k$ values. 
Note that the model's training signal only provides indirect information about execution accuracy, since the loss is computed against a single reference program (which ostensibly would execute correctly), not against a set of equivalent programs. 
These results are promising: while the T5-base model is very poorly-calibrated according to exact match, it is not nearly as bad when considering test-suite execution accuracy, which has been demonstrated to be the more informative metric \citep{zhong.r.2020}. 

\subsection{Easy and Hard TOD Subsets} \label{sec:easy_hard}
\cref{fig:all_models} shows that for many of the TOD models we examine, token-level confidence is well-correlated with accuracy; in \cref{tab:seq_level}, using {\tt{min}} to aggregate token-level confidence scores, we can obtain relatively low ECE at the sequence-level as well, meaning we can predict (on average) how likely a test example is to be correct.
The ECE values in \cref{tab:seq_level} tell us that the lower a model's {\tt{min}} confidence is, the more likely it is to make a mistake. 
Since different models have qualitatively different calibration curves, and may have complementary errors, we use an ensemble of three different models (MISO, T5-large, BART-large) to extract \textsc{easy} and \textsc{hard} splits of high- and low-confidence programs. 
These splits follow in the spirit of adversarial filtering \citep{zellers.r.2018} but are based on confidence rather than accuracy.

The \textsc{hard} subset contains the union of examples for which the \emph{sequence}-level confidence falls below a threshold -- all other examples are in \textsc{easy}. 
The threshold for each dataset is chosen by computing the $25^{th}$ percentile over all of the sequence-level probabilities across the three models.
This threshold is $0.86$ for SMCalFlow and $0.85$ for TreeDST. 
The union is taken across the three models, i.e. if any one model assigns an example a confidence below the threshold, the example is considered hard.

\cref{tab:hard_perc} shows the percentage of test data below the threshold for each model. 
Note that, because the models do not all assign low confidence to the same examples, the union of examples exceeds \perc{25} of the data.
Similarly, because the threshold is computed using the aggregated data from across all 3 models, it is possible for a single model to have more than \perc{25} \textsc{hard} examples, as long as the average across all 3 models is \perc{25}. 
For SMCalFlow, MISO contributes the most \textsc{hard} examples, while for TreeDST, BART-large contributes more. 

\begin{table}[ht]
    \centering
    \begin{tabular}{l c c }
    \hline\hline
     Model &\textsc{hard} (SM) & \textsc{hard} (Tr) \\
    \hline\hline
     MISO &  \perc{33.31}  & \perc{18.58}  \\
     BART-large & \perc{18.54}  & \perc{44.50} \\
     T5-large & \perc{23.15} & \perc{11.93} \\
     \hline
     Union & \perc{40.38} & \perc{50.50} \\
     \hline
    \end{tabular}
    \caption{Percentage of test examples labeled as \textsc{hard} by each model for SMCalFlow (SM) and TreeDST (Tr).}
    \label{tab:hard_perc}
\end{table}

\cref{tab:easy_vs_hard} shows the accuracy of each model on our subsets. 
We see much lower performance across all models on the \textsc{hard} subset and much higher performance on \textsc{easy}. 
MISO's performance is lower than that of the other models; the difference is larger than the performance difference in \cref{fig:all_models}. 
This is partly due to MISO contributing a large percentage of low-confidence examples to \textsc{hard} (cf. \cref{tab:hard_perc}) -- low-confidence examples are often more likely to be misclassified \citep{hendrycks.d.2016}.
We release our \textsc{hard} and \textsc{easy} subsets for both SMCalFlow and TreeDST, to act as challenge datasets for future work. 
\begin{table}[ht]
    \centering
    \begin{tabular}{l l c c}
    \hline\hline
     Dataset & Model & \textsc{hard} & \textsc{easy} \\
     \hline\hline
     \multirow{3}{*}{SMCalFlow} & MISO & 53.43 & 96.05 \\
     & BART-L & 62.66 & 96.15 \\
     & T5-L &  60.28 & 96.25 \\
     \hline
     \multirow{3}{*}{TreeDST} & MISO & 80.32 & 94.42 \\
     & BART-L & 84.97 & 98.67 \\
     & T5-L & 84.10 & 98.61 \\
     \hline
    \end{tabular}
    \caption{Exact match accuracy on the \textsc{easy} and \textsc{hard} subsets for all models. 
    L indicates ``large'' variant.  
    All models perform significantly worse on \textsc{hard}.}
    \label{tab:easy_vs_hard}
\end{table}

\vspace{-0.5em}
\subsection{Limitations}
Our study is limited by the models, datasets, and languages we consider.
Firstly, we examine only English datasets, limiting the impact of our results; future work may examine calibration across typologically diverse languages. 
Additionally, although we consider multiple datasets and models, our datasets are drawn from two domains and programming paradigms, and our models are limited to Transformer-based architectures.
While these choices are representative of current standards in executable semantic parsing, we encourage broader investigations spanning additional models and datasets. 
Specifically, future work may address programs that are interleaved into text-based interactions.
As sequence-models become integrated into chat-bot interfaces, increasing attention has been paid to augmenting these models with ``tools'', i.e. the ability to call APIs by predicting short programs \citep{schick.t.2023,mialon.g.2023}.
Examining calibration in these settings is particularly relevant. 
We examine only the highest-resource settings and leave examining how calibration profiles change with dataset size to future work.

We are also limited in how we measure calibration. 
ECE can be a brittle metric \citep{ovadia.y.2019, si.c.2022a}; we have tried to mitigate this by using adaptive binning. 
However, there are still hyperparameters and design choices involved in measuring ECE, and raw ECE scores can obscure a model's true calibration characteristics. 
We attempt to balance this using qualitative assessments. 
Similarly, while accuracy is easier to measure in semantic parsing than other text generation tasks, we have noted some of the shortcomings of exact-match accuracy metrics. 

\section{Conclusion}
Broadly, our results show that calibration is a complex phenomenon with a multitude of influences. 
On a single dataset, different models vary in their calibration error; moreover, the same model often varies drastically between different datasets, indicating that calibration is a function of both the model and the dataset. 
Taken together, these results point to the complexity of measuring calibration, and suggest that care should be taken when making claims about classes of models based on evidence from a limited number of models or datasets. 
Given the utility of calibrated models and the importance of calibration to safety, we advocate for considering calibration in the standard evaluation suite for semantic parsing models, and release our metric and visualization suite as a standalone package to facilitate these comparisons.

\section{Acknowledgements}
We would like to thank Zhengping Jiang, Anthony Platanios, Chenglei Si, Subhro Roy, Kate Sanders, Yu Su, Dan Klein, Matt Gardner, Anqi Liu, and Daniel Khashabi for their feedback and pointers.
We also thank the TACL reviewers and the Action Editor, who provided valuable feedback.
Elias Stengel-Eskin is supported by an NSF GRFP, and this work was additionally supported by NSF $\#1749025$.

\bibliography{calibration}
\bibliographystyle{acl_natbib}

\end{document}